\documentclass{scrartcl}
\pdfoutput=1
\usepackage[T1]{fontenc}
\usepackage{lmodern}
\usepackage{graphicx}
\usepackage[applemac]{inputenc}
\usepackage{ngerman}
\usepackage{hyperref}
\usepackage{tikz}
\usepackage{pgf}

\usepackage{cite}
\usepackage{graphicx}
\usepackage{subfigure}
\usepackage{amssymb}
\usepackage{amsmath}
\usepackage{amscd}
\usepackage{amsfonts}
\usepackage{amsthm}

\newcommand{\commentout}[1]{}

\newtheorem{theorem}{Theorem}

\newtheorem{proposition}{Proposition}
\newtheorem{corollary}{Corollary}
\newtheorem{example}{Example}

\newcommand{\proofskip}{\vspace{3ex}}
\newcommand{\introskip}{\vspace{3ex}}

\newcommand{\Proof}{\noindent \textbf{Proof:}\quad}

\newcommand{\R}{\mathbb{R}}                    

\newcommand{\AG}[1]{\mathcal{G}_{#1}}
\newcommand{\AGA}{\AG{\!\mathcal{A}}}

\DeclareMathOperator{\id}{id}                  
\DeclareMathOperator{\typeOf}{type}            

\newcommand{\abb}[1]{#1}                       
\renewcommand{\S}[1]{{\mathcal{#1}}}           
\def\vec#1{\mathchoice{\mbox{\boldmath$\displaystyle#1$}}
{\mbox{\boldmath$\textstyle#1$}}
{\mbox{\boldmath$\scriptstyle#1$}}
{\mbox{\boldmath$\scriptscriptstyle#1$}}}

\newcommand{\widebar}[1]{\overline{\!#1}}   

\newenvironment{case}{%
\left\{\begin{array}{c@{\quad : \quad}l}}%
{%
\end{array}\right.}

\begin{document}

\title{A Necessary and Sufficient Condition for Graph Matching Being
  Equivalent to the Maximum Weight Clique Problem}
\author{Brijnesh J.~Jain and Klaus Obermayer\\
Berlin University of Technology, Germany\\
\texttt{\{jbj|oby\}@cs.tu-berlin.de}}

\maketitle

\begin{abstract}
  This paper formulates a necessary and sufficient condition for a generic
  graph matching problem to be equivalent to the maximum vertex and edge
  weight clique problem in a derived association graph. The consequences of
  this results are threefold: first, the condition is general enough to
  cover a broad range of practical graph matching problems; second, a proof
  to establish equivalence between graph matching and clique search reduces
  to showing that a given graph matching problem satisfies the proposed
  condition; and third, the result sets the scene for generic continuous
  solutions for a broad range of graph matching problems. To illustrate the
  mathematical framework, we apply it to a number of graph matching
  problems, including the problem of determining the graph edit distance.
\end{abstract}

\section{Introduction}

The poor representational capabilities of feature vectors
triggered the field of \emph{structural pattern recognition} in the early
$1970$s to focus on pattern analysis tasks, where structured data is
represented in terms of strings, trees, or graphs.  A key issue in
structural pattern recognition is to measure the proximity of two structural
descriptions in terms of their structurally consistent or inconsistent parts. 
There is also strong evidence that human cognitive models of comparison and analogy establish relational
correspondences between structured objects \cite{Gentner83, Johnson88,
  Goldstone94}. The problem of measuring the structural proximity of graphs,
more generally referred to as the \emph{graph matching problem}, is often
computationally inefficient. Therefore, an appropriate formulation of the
graph matching problem is important to devise efficient solutions and to
gain insight into the nature of the problem.

One popular technique is to transform graph matching to an equivalent clique
search in a derived auxiliary structure, called \emph{association graph}
\cite{Ambler73, Barrow76,Pelillo99a,Pelillo99b,Raymond02}. Chen \& Yun \cite{Chen98} 
generalized association graph techniques by compiling results from 
\cite{Barrow76, Chen96, Kann92} and showing that the maximum common (induced) 
subgraph problem and its derivations can be  casted to a maximum clique problem. 
Pelillo \cite{Pelillo02} extended this collection by transforming the problem of matching free
 trees\footnote{A free tree is a directed acyclic graph without a \emph{root}.} to the \abb{MVCP} so
  that connectivity is preserved. Bunke \cite{Bunke97} showed that for
  special cost functions, the error correcting graph matching problem and
  the maximum common subgraph are equivalent. As a consequence, special
  graph edit distances can be computed via clique search in an association
  graph. Sch\"adler \& Wysotzki \cite{Schaedler99a} mapped the best
  monomorphic graph matching problem to a maximum weighted clique problem 
  without presenting a sound theoretical justification of their transformation.

Although association graph techniques are attractive, an equivalence
relationship between graph matching and clique search has been established
only for certain classes of graph matching problems. Examples for graph
matching problems, which have been proven equivalent to clique search in an
association graph are the maximum common (induced) subgraph and special
cases thereof, or the graph and tree edit distance for special cost
functions. To utilize the benefits of association graph techniques for a possibly 
broad range of graph matching problems, we pose the following question:

\begin{quote}
  \emph{Under which conditions is graph matching equivalent to clique search
    in an association graph?}
\end{quote}

\smallskip

\noindent
With the above question in mind, we make the following contributions:
\begin{itemize}
\item We formulate a necessary and sufficient condition (C), for graph
  matching problems to be equivalent to the maximum weight clique problem in
  an association graph.  As opposed to standard formulations, the maximum
  weight clique problem takes into account weights assigned to both vertices
  and edges.
\item Condition (C) is sufficiently strict to cover a broad range of graph
  matching problems.  Besides the well-known standard problems, we show, for
  example, that determining the graph edit distance is equivalent to the
  maximum weight clique problem in an association graph.
\item The mathematical framework developed in this paper provides a proof
  technique that reduces the problem of showing equivalence between graph
  matching and clique search to the problem of showing that a given graph
  matching problem satisfies condition (C). By means of a number of
  examples, we illustrate how the novel technique considerably simplifies
  equivalence proofs.
\end{itemize}

The paper is organized as follows. We conclude this section by introducing
the terminology. Section 2 presents the background and an
intuitive idea of condition (C). In Section 3, we formulate
condition (C) and prove that (C) is necessary and sufficient for the desired
equivalence realtionship. We apply (C) in Section 4.
Finally, Section \ref{sec:conclusion} concludes with a summary of the main
results and an outlook on further research.

\subsection{Preliminaries}

The aim of this subsection is to introduce the terminology and notations
used througout this contribution. We assume knowledge about basic graph
theory. 

\subsection*{Basic Graph Theory}

\paragraph*{Graphs}
For convenience of presentation, all graphs are undirected without loops.
The graphs we consider are triples $X = (V, E, \vec{X})$ consisting of a
finite set $V = V(X)$ of vertices, a set $E = E(X)$ of edges, and an
\emph{attributed adjacency matrix} $\vec{X} = (x_{ij})$ with elements
$x_{ij}$ from a set of \emph{attributes} $\S{A} \cup\{\epsilon\}$. Elements
$x_{ij} \in \S{A}$ are attributes either assigned to vertices if $i = j$ or
to edges if $(i,j) \in E$. We label non-edges $(i,j) \in \widebar{E}$ with a
distinguished \emph{void attribute} $x_{ij} = \epsilon$.

\smallskip

\paragraph*{Subgraphs}
We write $X'\subseteq X$ to denote that $X'$ is a subgraph of $X$. For a
subset $U$ of $V(X)$, the graph $X[U]$ denotes the induced subgraph of $X$
induced by $U$. 

\smallskip

\paragraph*{Items}
To unclutter the text from tedious case distinctions, we occasionally make
use of the notion of \emph{item}. Items of $X$ are elements from $I(X) = V
\times V$. Thus, an item $\vec{i} = (i, j)$ of $X$ is either a vertex if $i
= j$, an edge if $\vec{i} \in E(X)$, or a non-edge if $\vec{i} \in
\widebar{E}(X)$.

\smallskip

\paragraph*{Morphisms}
Let $X$ and $Y$ be graphs. A \emph{morphism} from $X$ to $Y$ is a mapping
\[
\phi: V(X) \rightarrow V(Y), \quad i \mapsto i^{\phi}.
\]
A \emph{partial morphism} from $X$ to $Y$ is a morphism $\phi$ defined on a
subset of $V(X)$. By $\S{D}(\phi)$ we denote the \emph{domain} and by
$\S{R}(\phi)$ the \emph{range} of a partial morphism $\phi$.  By abuse of
notation we occasionally write $\vec{i} = (i,j) \in \S{D}(\phi)$ if $i \in
\S{D}(\phi)$ and $j \in \S{D}(\phi)$. The meaning of $\vec{i}\in
\S{R}(\phi)$ is now obvious. A \emph{monomorphism} is an injective morphism.
We use the notions \emph{homo-} and \emph{isomorphism} as in standard graph
theory. A partial \emph{subgraph-morphism} is an isomorphism between
subgraphs.

\subsection*{The Maximum Weight Clique Problem}\label{sec:MWCP}

Let $X = (V, E, \vec{X})$ be a graph with attributes from $\S{A} = \R
\cup\{\epsilon\}$.  A \emph{clique} of $X$ is a subset $C \subseteq V$ such
that the induced subgraph $X[C]$ is complete. A clique is said to be
\emph{maximal} if $C$ is not contained in any larger clique of $X$, and
\emph{maximum} if $C$ has maximum cardinality of vertices. The \emph{weight}
$\omega(C)$ of a clique $C$ of $X$ is defined by
\begin{equation}\label{eq:MWCP:weight_of_clique}
\omega(C) = \sum_{i,j \in C} x_{ij}.
\end{equation}
The weight of a clique $C$ is the total of all vertex and edge weights of
the subgraph $X[C]$ induced by $C$.  Since the vertices of $X[C]$ are
mutually adjacent, the void symbol $\epsilon$ does not occur in the sum of
(\ref{eq:MWCP:weight_of_clique}). Hence, $\omega(C)$ is well-defined.

The \emph{maximum weight clique problem} is a combinatorial optimization
problem of the form
\begin{equation}\label{nlp:MWCP:QIP:mwcp}
\mbox{$
  \begin{array}{l@{\qquad}l}
    \mbox{maximize}& {\displaystyle \omega(C) = \sum_{i,j \in C} x_{ij}}
    \\[1ex]
    \mbox{subject to} & {\displaystyle C \in \S{C}_{X}},
  \end{array}
  $}
\end{equation}
where $\S{C}_{X}$ is the set of all cliques of $X$.

A \emph{maximal weight clique} of $X$ is a clique $C\in\S{C}_{X}$ such that
\[
C \subseteq C' \;\Rightarrow\;\omega(C) \geq \omega(C')
\]
for all cliques $C'$ of $X$. It is impossible to enlarge a maximal weight
clique $C$ to a clique with higher weight. If all vertices and edges of $X$
are associated with positive weights, a maximal weight clique is not a
proper subset of another clique. A \emph{maximum weight clique} of $X$ is a
clique $C\in \S{C}_{X}$ with maximum total weight over its vertices and
edges. By $\S{C}_{X}^{\times}$ we denote the set of all maximal cliques and
by $\S{C}_{X}^{*}$ the set of all maximum cliques of $X$.

\section{Background}

The aim of this section is twofold: First, it introduces the problem and
motivates its solution. To this end, we consider the classical maximum
common induced subgraph problem (\abb{MCISP}). Second, it provides an
intuitive idea of how to solve that problem in a more general setting. 

\subsection{The Problem}

To set the scene, we consider the \abb{MCISP}. Given two graphs $X$ and $Y$,
the \emph{\abb{MCISP}} asks for a partial isomorphism $\phi: V(X)
\rightarrow V(Y)$ that maximizes the cardinality $|\S{D}(\phi)|$ of its
domain. This optimization problem is called \abb{MCISP}, because each
partial isomorphism $\phi$ between $X$ and $Y$ is an isomorphism between the
induced subgraphs $X[\S{D}(\phi)]$ and $Y[\S{R}(\phi)]$. Since maximizing
$|\S{D}(\phi)|^{2}$ instead of $|\S{D}(\phi)|$ does not effect the problem,
we may rewrite the \abb{MCISP} as

\begin{equation}\label{nlp:GMP:MCISP}
\mbox{$
  \begin{array}{l@{\qquad}l}
    \mbox{maximize}& {\displaystyle f\big(\phi, X, Y\big) = |\S{D}(\phi)|^{2} 
        = \sum_{i,j\in \S{D}(\phi)}\kappa_{ij\,i^{\phi}j^{\phi}}}
    \\[1ex]
    \mbox{subject to}
    & {\displaystyle \phi \in \S{M}},
  \end{array}
  $}
\end{equation}
where the \emph{search space} $\S{M}$ is the set of all partial isomorphisms
from $X$ to $Y$. The values $\kappa_{ijrs} \in \{0, 1\}$ are the
\emph{compatibility values} with
\begin{equation}\label{eq:GMP:Exact_GMP_kappa}
\kappa_{ijrs} =
\begin{case}
  1& x_{ij} = y_{rs}\\
  0 & \mbox{otherwise}
\end{case}
\end{equation}
for all $i,j \in V(X)$ and $r, s \in V(Y)$. A compatibility value
$\kappa_{ijrs}$ indicates whether item $(i,j)$ of $X$ has the same attribute
as item $(r,s)$ of $Y$. The \emph{matching objective} $f$ counts the number
of items from $X$ consistently mapped to their exact counterparts in $Y$.

It is well known that the \abb{MCISP} is \abb{NP}-complete \cite{Garey79}.
Therefore, exact algorithms that guarantee to return an optimal solution are
useless for all but the smallest graphs. In a practical setting, the time
required to compute an optimal solution will typically reduce the overall
\emph{utility} of the algorthim. It is rather more desirable to trade
quality with time and to provide near-optimal solutions within an acceptable
time limit. Thus, it is conducive to consider local optimal solutions of
(\ref{nlp:GMP:MCISP}).  We say a partial isomorphism $\phi \in \S{M}$ is a
local optimal solution if there is no other partial isomorphism from $\S{M}$
with larger domain.

To solve the maximum common induced subgraph problem, we transform it to
another combinatorial optimization problem using association graph
techniques originally introduced by \cite{Ambler73, Levi72, Barrow76}. An
association graph $Z = X \diamond Y$ of $X$ and $Y$ is a graph with vertex
and edge set
\begin{align*}
  V(Z) &= \big\{ir \,:\, x_{ii} = y_{rr} \big\} \subseteq V(X) \times V(Y) \\
  E(Z) &= \big\{(ir, js) \,:\, x_{ij} = y_{rs} \big\} \subseteq V(Z) \times
  V(Z).
\end{align*}
The attributed adjacency matrix $\vec{Z} = (z_{irjs})$ of $Z$ is defined by
\[
z_{irjs} = \begin{case}
  1 & ir = js \mbox{ and } ir\in V(Z) \\
  1 & ir \neq js \mbox{ and } (ij, rs) \in E(Z) \\
  \epsilon & \mbox{otherwise}
\end{case}
\]
for all $ir, js \in V(Z)$. Thus, we assign to all vertices and edges of $Z$
the weight $1$ and to all non-edges the void attribute $\epsilon$. Note that
we derived the weights $z_{irjs}$ by inserting the corresponding
compatibility values $\kappa_{ijrs}$.

By definition of an association graph, we have the following useful
equivalence relationship between between the partial isomorphisms from
$\S{M}$ and the cliques in $Z$:
\begin{itemize}
\item $Z$ uniquely encodes each partial isomorphism $\phi$ from $\S{M}$ as a
  clique $C_{\phi}$ in $Z$ such that $|\S{D}(\phi)| = |C|$.
\item The set $\S{C}_{Z}(\S{M})$ of cliques encoding partial isomorphisms is
  equal to the set $\S{C}_{Z}$ of all cliques in $Z$.
\end{itemize}
The above equivalence directly implies that the maximum (maximal) cliques of
$Z$ are in one-to-one correspondence with the global (local) optimal partial
isomorphisms from $\S{M}$. Hence, solving the \abb{MCISP}
(\ref{nlp:GMP:MCISP}) is equivalent to solving the maximum clique problem in
$Z$.

\bigskip

\noindent
\emph{Benefits of the Association Graph Framework}

\smallskip

What makes an association graph formulation of the \abb{MCISP} so useful is
that
\begin{itemize}
\item the maximum clique problem is mathematically well founded,
\item it provides us access to a plethora of clique algorithms to solve the
  original problem,
\item abstracts from the particularities of the graphs being matched, and
\item abstracts from the constraints on feasible matches.
\end{itemize}

The benefits of the second item on clique algorithms is worth to be
discussed in more detail. Both the \abb{MCISP} and the maximum clique
problem are combinatorial problems. Solution techniques for combinatorial
problems can be classified into two groups:
\begin{itemize}
\item \emph{Discrete methods} generate a sequence of suboptimal or partial
  solutions to the original problem. This process is guided by local search
  combined with techniques to escape from local optimal solutions.
\item \emph{Continuous methods} embed the discrete solution space in a
  larger continuous one.  Exploiting the topological and geometric
  properties of the continuous space, the algorithm creates a sequence of
  points that converges to the solution of the original problem.
\end{itemize}
Discrete approaches suffer from problems having a large number of local
maxima. On the other hand, continuous approaches such as interior point
methods reveal to be efficient solutions to large scale combinatorial
optimization problems. Hence, using association graph techniques provides us
access to those efficient continuous solutions.

\bigskip

\noindent
\emph{Shortcomings of the Association Graph Framework}

\smallskip

Despite its long tradition and benefits, the association graph formulation
still suffers from the following shortcomings:
\begin{itemize}
\item It is unclear which graph matching problems are equivalent to clique
  search in an association graph. The equivalence relationship in question
  has been proved only for selected problems. For example, Chen \& Yun
  \cite{Chen98} compiled results from \cite{Barrow76, Chen96, Kann92},
  showing that the maximum common (induced) subgraph problem and its
  derivations can be casted to a maximum clique problem.  Pelillo and his
  co-workers extended this collection for different types of tree matching
  problems \cite{Bartoli00, Pelillo99a, Pelillo99d, Pelillo02}.
\item In all these examples, equivalence proofs of graph matching and clique
  search follow the same recurring pattern: First construct an appropriate
  association graph and then establish a bijective mapping between cliques
  and morphisms. There is no mathematical framework that reduces equivalence
  proofs to showing that the preconditions of some generic equivalence
  relationship between graph matching and clique search are satisfied.
\end{itemize}

\bigskip

\noindent
\emph{What we want}

\smallskip

The aim is to remove both shortcomings addressed in the previous paragraph.
We want to formulate a necessary and sufficient condition (C) for
equivalence between graph matching and clique search. The existence of
condition (C) explains which graph matching problems are equivalent to
clique search in an association graph and reduces equivalence proofs to
showing that a given graph matching problem satisfies (C). In addition, we
want to indicate by a number of examples that a broad range of graph
matching problems satisfy condition (C) and are therefore equivalent to
clique search in an association graph.

\subsection{The Idea}

Our goal is to present an intuitive idea of a necessary and sufficient
condition (C) for equivalence between graph matching and clique search. To
this end, we turn from the \abb{MCISP} to the general case. Suppose that $X$
and $Y$ are two graphs, and let $\S{M}_{XY}$ be the set of all partial
morphisms from $X$ to $Y$. The graph matching problem considered here
generalizes the \abb{MCISP} in two ways:
\begin{itemize}
\item By allowing nonnegative real-valued compatibility values
  $\kappa_{ijrs} = \kappa_{jisr}$.
\item By allowing any subset of $\S{M}_{XY}$ as search space.
\end{itemize}
Allowing real-valued compatibility values $\kappa_{ijrs} \geq 0$ enables us
to measure the degree of \emph{compatibility}, or \emph{consistency},
between items $\vec{i} = (i,j)$ of $X$ and $\vec{r} = (r,s)$ of $Y$. This
generalization is useful to cope with noisy attributes. Considering
arbitrary subsets of $\S{M}_{XY}$ provides a flexible mechanism to cope with
structural errors.

Applying both generalizations, the \emph{graph matching problem} (\abb{GMP})
is a combinatorial optimization problem of the form
\begin{equation}\label{nlp:GMP:gmp}
\mbox{$
  \begin{array}{l@{\qquad}l}
    \mbox{maximize}& {\displaystyle f\big(\phi, X, Y\big) = 
      \sum_{\vec{i}\in\S{D}({\phi})}\kappa_{\vec{i}\vec{i}^{\phi}}}\\
    \mbox{subject to}
    & {\displaystyle \phi \in \S{M}},
  \end{array}
  $}
\end{equation}
where the search space $\S{M}$ is a subset of $\S{M}_{XY}$, and
$\vec{i}^{\phi} = (i^{\phi},j^{\phi}) \in I(Y)$ is the image item of item
$\vec{i} = (i,j) \in I(X)$. Problem (\ref{nlp:GMP:gmp}) is a generic
formulation that subsumes a broad range of practical graph matching problems.

Often, solving an instance of problem (\ref{nlp:GMP:gmp}) is computationally
intractable. As for the \abb{MCISP}, we therefore consider
simpler versions of (\ref{nlp:GMP:gmp}) that allow local optimal solutions.
A partial morphism $\phi \in \S{M}$ is a local optimal solution of
(\ref{nlp:GMP:gmp}) if $\S{D}(\phi) \subseteq \S{D}(\psi)$ implies $f(\phi,
X, Y) \geq f(\psi, X, Y)$ for all $\psi \in \S{M}$.

\bigskip

\noindent
\emph{Applying the Association Graph Framework}

\smallskip

The central question at issue is: Under which conditions is the generic
graph matching problem (\ref{nlp:GMP:gmp}) equivalent to clique search in a
derived association graph $Z = X \diamond Y$? Certainly, an association
graph $Z$ should satisfy the following properties:
\begin{description}
\item[P1.] $Z$ uniquely encodes each partial morphism $\phi$ from the search
  space $\S{M}$ as a clique $C_{\phi}$ in $Z$ such that $f(\phi, X, Y) =
  \omega(C)$, where $\omega(C)$ denotes the weight of clique $C$.
\item[P2.] The set $\S{C}_{Z}(\S{M}) = \{C_{\phi} \in \S{C}_{Z}
  \,:\,\phi\in\S{M}\}$ of all cliques encoding partial morphisms from
  $\S{M}$ is equal to the set $\S{C}_{Z}$ of all cliques in $Z$.
\end{description}

\medskip

\noindent
\emph{Ad P1:} We can always derive an association graph satisfying the first
property P1.  Consider the complete graph $Z_{XY}$ with vertex set $V_{XY} =
V(X) \times V(Y)$. To each item $\big((i,j),(r,s)\big) \in I(Z_{XY})$, we
assign the weight $\kappa_{ijrs}$. Since $Z_{XY}$ is complete, there are
only vertex and edge items. We extract a subgraph from $Z_{XY}$ to form an
association graph $Z$ as follows: For each partial morphism $\phi\in \S{M}$,
we construct a complete subgraph $Z_{\phi}$ of $Z_{XY}$ with vertex set
$V(Z_{\phi}) = C_{\phi} = \{ii^{\phi}\,:\, i \in \S{D}(\phi)\}$. The vertex
set $C_{\phi}$ is a clique in $Z_{XY}$ such that $f(\phi, X, Y) =
\omega(V_{\phi})$. We obtain an association graph $Z \subseteq Z_{XY}$ by
taking the union of all subgraphs $Z_{\phi}$ with $\phi \in\S{M}$. Then be
construction $Z$ satisfies property P1. Note that construction of $Z$ in the
above way is impractical, because it requires enumeration of all members of
$\S{M}$. But it provides a simple way to show that there is always an
association graph satisfying property P1.

\bigskip

\noindent
\emph{Ad P2:} Now let us turn to the second property P2. We first provide a
fictitious example to show that $P2$ does generally not hold.
\begin{example}\label{ex:idea:1}
  Consider the set $\S{M}_{2}$ of partial morphisms $\phi: V(X) \rightarrow
  V(Y)$ that are defined on subsets of exactly two vertices from $X$,
  i.e.~$\big|\S{D}(\phi)\big| = 2 < |X|$. Suppose that we construct $Z$ as
  described in the previous paragraph. Then the set $\S{C}_{Z}(\S{M}_{2})$
  of all cliques encoding morphisms from $\S{M}_{2}$ are cliques in $Z$ with
  exactly two vertices. We may encounter the following pitfalls:
  \begin{description}
  \item[PF1.] Let $C = \{i,j\}$ be a clique in $Z$ encoding
    $\phi\in\S{M}_{2}$.  Then $\{i\}$ and $\{j\}$ are both cliques in $Z$
    not contained in $\S{C}_{Z}(\S{M}_{2})$.
  \item[PF2.] Let $C, C', C''$ be cliques from $\S{C}_{Z}(\S{M}_{2})$ with
    $C = \{i,j\}$, $C' = \{j, k\}$, and $C'' = \{i,k\}$. Then $C =
    \{i,j,k\}$ is a clique in $Z$ with three vertices and therefore not
    contained in $\S{C}_{Z}(\S{M}_{2})$.
  \end{description}
  Hence, we have $\S{C}_{Z}(\S{M}_{2}) \neq \S{C}_{Z}$ and therefore a graph
  matching problem (\ref{nlp:GMP:gmp}) defined on $\S{M}_{2}$ is not
  equivalent to clique search in an association graph.
\end{example}
  
Example \ref{ex:idea:1} indicates that equivalence between graph matching
and clique search depends on the structure of the search space $\S{M}$.  So
our central question of issue reduces to a necessary and sufficient
condition on $\S{M}$ such that $\S{C}_{Z}(\S{M}) = \S{C}_{Z}$.

\bigskip

\noindent
\emph{A Necessary and Sufficient Condition}

\smallskip

The goal is to present an intuitive idea of a necessary and sufficient
condition for $\S{C}_{Z}(\S{M}) = \S{C}_{Z}$. Our claim is that
$\S{C}_{Z}(\S{M}) = \S{C}_{Z}$ holds whenever there is a \emph{property}
that completely describes the set $\S{M}$.

First, we specify the notion of property. Given a graph matching problem,
the purpose of a property is to describe the characteristics of the search
space $\S{M}$ in such a way that we can derive the desired equivalence
relationship. For example, let $\S{M}$ be the set of all partial morphisms
$\phi$ from $\S{M}_{XY}$ that satisfy the property $\mathfrak{p}$ to be
injective. Although property $\mathfrak{p}$ completely describes the
characteristics of $\S{M}$, it is not suitable to deal with it conveniently.
The reason is as follows: According to the pitfalls PF1 and PF2 of Example
\ref{ex:idea:1}, we have to show that restrictions and feasible unions of
morphisms that satisfy property $\mathfrak{p}$ also satisfy $\mathfrak{p}$.
This task is unnecessarily complex and can be simplified by imposing a
locality restriction on the notion of property.

A \emph{property} $\mathfrak{p}$ on a set $\S{S}$ is a binary function
$\mathfrak{p}:\S{S} \rightarrow \{0,1\}$. We say an element $x \in \S{S}$
\emph{satisfies} property $\mathfrak{p}$ if $\mathfrak{p}(x) = 1$ for each
$x \in \S{S}'$.  We can extend the notion of a property to the subsets of
$\S{S}$ by considering their local behavior. A subset $\S{S}' \subseteq
\S{S}$ satisfies property $\mathfrak{p}$ if $\mathfrak{p}(x) = 1$ for each
element $x \in \S{S}'$, i.e.~if $\S{S}'$ locally satisfies $\mathfrak{p}$.

Next, we characterize an association graph $Z = X \diamond Y$ by means of a
property. To this end, we consider the complete graph $Z_{XY}$ defined on
the vertex set $V_{XY} = V(X)\times V(Y)$. Let $\mathfrak{p}$ be a property
on the set $I_{XY} = I(Z_{XY})$ of items such that the vertices and edges of
$Z$ are all the items that satisfy $\mathfrak{p}$. Hence, the subsets of
$I_{XY}$ that satisfy property $\mathfrak{p}$ are the induced subgraphs of
$Z$ induced by its cliques. We can express the set of all cliques in $Z$ as
$\S{C}_{Z} = \{U \subseteq I_{XY} \,:\, U \mbox{ satisfies }
\mathfrak{p}\}$. Now using $\mathfrak{p}$, we rewrite condition
$\S{C}_{Z}(\S{M}) = \S{C}_{Z}$ as follows: \emph{There is a property
  $\mathfrak{p}$ such that $\S{C}_{Z}(\S{M}) = \{C \in \S{C}_{Z} \,:\, C
  \mbox{ satisfies } \mathfrak{p}\}$.} Since the cliques from
$\S{C}_{Z}(\S{M})$ encode the feasible morphisms from $\S{M}$, we may
directly translate the last condition to a condition on the search space
$\S{M}$:
\begin{description}
  \em
\item[(C)] There is a property $\mathfrak{p}$ such that $\S{M} = \{\phi \in
  \S{M}_{XY} \,:\, \phi \mbox{ satisfies } \mathfrak{p}\}$.
\end{description}
To summarize, we claim that condition (C) is necessary and sufficient for
the equivalence relationship in question.

\section{A Necessary and Sufficient Condition} \label{sec:GMP2MWCP}

The aim of this section is to prove that condition (C) is necessary and
sufficient for equivalence between graph matching and clique search in an
association graph.

\subsection{Closed Sets of $\mathfrak{p}$-Morphisms}

Here, we consider properties $\mathfrak{p}$ defined on the set $I_{XY} =
I(X) \times I(Y)$ of pairs of items from $X$ and $Y$. Let $(\vec{i},
\vec{r}) \in I_{XY}$. We say, items $\vec{i}$ and $\vec{r}$ are
\emph{$\mathfrak{p}$-similar}, written as $\vec{i}\sim_{\mathfrak{p}}
\vec{r}$, if $(\vec{i},\vec{r})$ satisfies property $\mathfrak{p}$. Let us
consider some examples.

\begin{example}\label{example:properties}
  Let $(\vec{i},\vec{r}) \in I_{XY}$, and let all morphisms $\phi$ be partial
  morphisms from $\S{M}_{XY}$. Then the following examples are properties on
  $I_{XY}$:
\begin{enumerate}
\item $\vec{i}\sim_{\mathfrak{p}} \vec{r} = $ there is a partial (mono-,
  subgraph-, homo-, iso-) morphism $\phi$ with $ \vec{i}^{\phi} = \vec{j}$.
\item $\vec{i}\sim_{\mathfrak{p}} \vec{r} = $ there is a partial
  connectivity preserving isomorphism $\phi$ with $\vec{i}^{\phi} =
  \vec{j}$.
\end{enumerate}
\end{example}
Next, we apply the notion of property on $I_{XY} = I(X) \times I(Y)$ to
partial morphisms. To this end, we identify a partial morphism $\phi\in
\S{M}_{XY}$ with the binary relation
\[
\Gamma(\phi) = \big\{(\vec{i},\vec{r}) \in \S{D}(\phi) \times \S{R}(\phi)
\,:\, \vec{i}^{\phi} = \vec{r}\big\} \subseteq I_{XY}.
\]  
We say, $\phi$ is a \emph{$\mathfrak{p}$-morphism} if $\Gamma(\phi)$ is a
subset of the set $\S{R}_{XY}^{\mathfrak{p}} = \big\{(\vec{i},\vec{r}) \in
I_{XY} \,:\,\vec{i}\sim_{\mathfrak{p}} \vec{r}\big\}$. In informal terms, a
partial morphism $\phi$ satisfies property $\mathfrak{p}$ if it locally
satisfies $\mathfrak{p}$. To illustrate the defintion of
$\mathfrak{p}$-morphism, we provide some examples. 
\begin{example}\label{example:p-morphisms}
  Consider the properties $\mathfrak{p}$ of Example
  \ref{example:properties}.
\begin{enumerate}
\item Let $\mathfrak{p}$ be the property of Example
  \ref{example:properties}.1, and let $\phi \in \S{M}_{XY}$ be a partial
  morphism. Since $\vec{i}^{\phi} = \vec{r}$ for all pairs
  $(\vec{i},\vec{r}) \in \Gamma(\phi)$, we trivially have
  $\vec{i}\sim_{\mathfrak{p}} \vec{r}$ for all pairs $(\vec{i},\vec{r}) \in
  \Gamma(\phi)$. Hence, $\Gamma(\phi) \subseteq \S{R}_{XY}^{\mathfrak{p}}$
  and therefore $\phi$ is a $\mathfrak{p}$-morphism. Similarly, mono-, homo,
  and isomorphisms are $\mathfrak{p}$-morphisms.
\item Let $\mathfrak{p}$ be the property of Example
  \ref{example:properties}.2, and let $\phi$ be a partial connectivity
  preserving morphism on $V(X)$. As in (1), we find that $\phi$ is a
  $\mathfrak{p}$-morphism. Note that for non-edge items $(i, j) \in
  \widebar{E}(X)$ the restriction $\phi'$ of $\phi$ to $\{i, j\}$ does not
  preserve connectivity. But what is important is that $\phi'$ can be
  extended to a partial morphism $\phi$ that preserves connectivity.
\end{enumerate}
\end{example}
Let $\S{M}_{XY}^{\mathfrak{p}} \subseteq \S{M}_{XY}$ denote the set of all
$\mathfrak{p}$-morphisms. We say, a subset $\S{M}$ of $\S{M}_{XY}$ is
\emph{$\mathfrak{p}$-closed} if $\S{M} = \S{M}_{XY}^{\mathfrak{p}}$. The
next result provides basic $\mathfrak{p}$-closed sets.

\begin{proposition}\label{prop:MWCP:gmp2mwcp:p-morphisms}
  The following subsets of $\S{M}_{XY}$ are $\mathfrak{p}$-closed:
\begin{enumerate}
\item $\S{M}=$ set of all partial morphisms
\item $\S{M}=$ set of all partial monomorphisms
\item $\S{M}=$ set of all partial homomorphisms
\item $\S{M}=$ set of all partial isomorphisms
\item $\S{M}=$ set of all partial subgraph-morphisms
\end{enumerate}
\end{proposition}

\Proof We only show the assertion for the set $\S{M}$ of all partial
isomorphisms . The proofs for the other sets are similar. Let $X$ and $Y$ be
graphs with adjacency matrices $\vec{X} = (x_{ij})$ and $\vec{Y} =
(y_{rs})$. We define a property $\mathfrak{p}$ on $I_{XY}$ such that
\[
\S{R}_{XY}^{\mathfrak{p}} = \big\{(\vec{i}, \vec{r}) \,:\, x_{\vec{i}} =
y_{\vec{r}}, \typeOf(\vec{i}) = \typeOf(\vec{r}) \big\} \subseteq I(X)
\times I(Y),
\]
where $\typeOf(\vec{i})$ maps item $\vec{i}$ to its type (vertex, edge, or
non-edge).

Next, we show that $\S{M} = \S{M}_{XY}^{\mathfrak{p}}$. Let $\phi$ be a
partial isomorphism from $X$ to $Y$, let $\vec{i}$ be an item of $X$, and
let $\vec{i}^{\phi} = \vec{r}$ be the image of $\vec{i}$ in $Y$. Since
$\phi$ is a partial isomorphism, we have $x_{\vec{i}} = y_{\vec{j}}$. In
addition, $\phi$ preserves the type of items $\vec{i}$ and $\vec{r}$. Hence,
the set $\Gamma(\phi)$ is a subset of $\S{R}_{XY}^{\mathfrak{p}}$. This
proves $\S{M} \subseteq \S{M}_{XY}^{\mathfrak{p}}$.

Now assume that $\phi$ is a $\mathfrak{p}$-morphism from
$\S{M}_{XY}^{\mathfrak{p}}$. Since $\Gamma(\phi)$ is a subset of
$\S{R}_{XY}^{\mathfrak{p}}$, it is sufficient to show that $\phi$ is
bijective. Assume that $\phi$ is not bijective. Then $\phi$ is not
injective, because $\phi$ is a partial morphism. Hence, there are distinct
vertices $i, j$ of $X$ with $i^{\phi} = j^{\phi} = r$. Then $\phi$ maps a
non-vertex item $(i,j)$ to a vertex item $(r, r)$, which contradicts the
condition $\typeOf(\vec{i}) = \typeOf(\vec{r})$. This proves
$\S{M}_{XY}^{\mathfrak{p}}\subseteq \S{M}$. Combining both results yields
the assertion. \qed

\proofskip

Though the proof of Proposition \ref{prop:MWCP:gmp2mwcp:p-morphisms} is
fairly simple, it illustrates the basic approach how to show that a certain
subset $\S{M}$ of $\S{M}_{XY}$ is $\mathfrak{p}$-closed. The task is to
construct an appropriate property $\mathfrak{p}$ such that we can show
$\S{M} = \S{M}_{XY}^{\mathfrak{p}}$. 

We conclude this section with a formal proof that the set $\S{M}_{2}$ from
Example \ref{ex:MWCP:gmp2mwcp:not-closed} is not $\mathfrak{p}$-closed.

\begin{example}\label{ex:MWCP:gmp2mwcp:not-closed}
  Consider the set $\S{M}_{m}$ of partial morphisms $\phi: V(X) \rightarrow
  V(Y)$ with $\big|\S{D}(\phi)\big| \leq m < |X|$, where $m\geq 2$. There is
  no property $\mathfrak{p}$ on $I_{XY}$ such that $\S{M}_{m}$ is
  $\mathfrak{p}$-closed.
\end{example}
\Proof Let $\mathfrak{p}$ denote the property on $I_{XY}$ such that the set
of all partial morphisms $\S{M}_{XY}$ is $\mathfrak{p}$-closed. According to
Proposition \ref{prop:MWCP:gmp2mwcp:p-morphisms}, such a property exists.
Consider the set $\S{M}_{2}$. It is easy to see that
\[
\bigcup_{\phi\in\S{M}_{2}} \Gamma(\phi) = \S{R}^{\mathfrak{p}}_{XY}.
\]
On one hand, the relation $\S{R}^{\mathfrak{p}}_{XY}$ is too large, because
it admits arbitrary partial morphisms as $\mathfrak{p}$-morphisms. On the
other hand, $\S{R}^{\mathfrak{p}}_{XY}$ is a minimal set in the following
sense: If we remove an element $(\vec{i}, \vec{r})$ from
$\S{R}^{\mathfrak{p}}_{XY}$, then the morphism $\phi\in \S{M}_{2}$ with
$\vec{i}^{\phi} = \vec{r}$ is no longer a $\mathfrak{p}$-morphism. This
shows the assertion.  \qed

\begin{figure}[tbp]
  \centering \includegraphics[width=12cm]{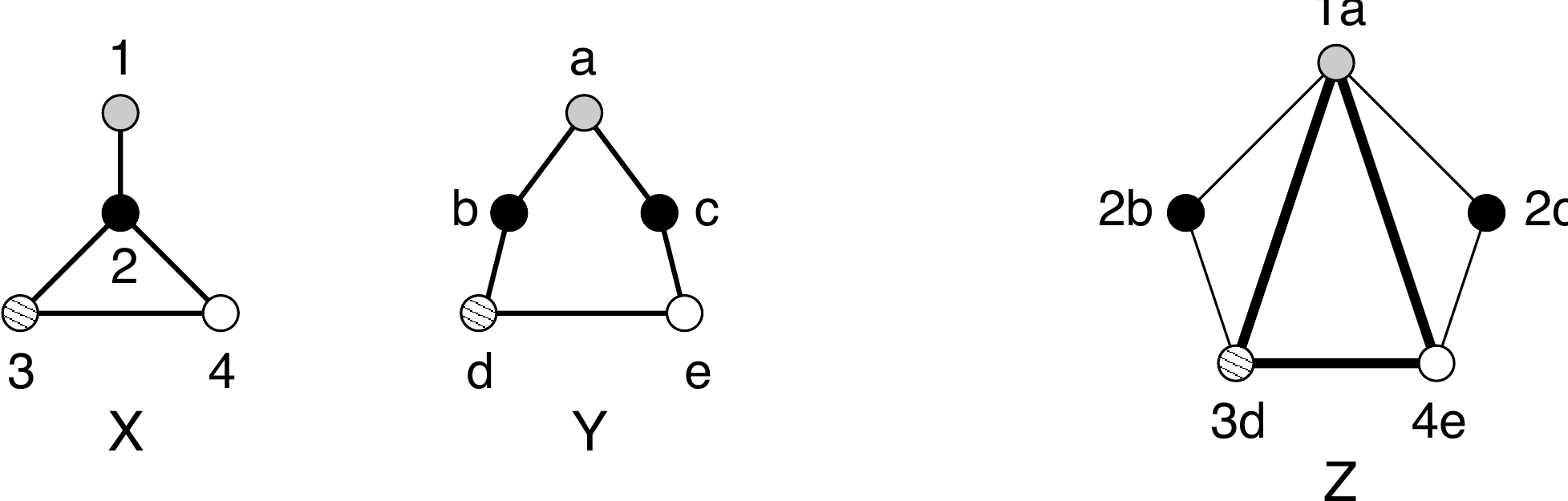}
  \caption{Graphs $X$ and $Y$ with association graph $Z = X
    \diamond Y$ for the maximum common connectivity preserving induced
    subgraph problem. Different fillings refer to different vertex
    attributes. All edges have the same attribute. Highlighted edges in $X$
    and $Y$ correspond to isomophic subgraphs. Note that the highlighted
    subgraph in $X$ is not induced. Highlighted edges in $Z$ refer to the
    maximum clique. }
  \label{fig:example:connectivity}
\end{figure}
\subsection{Construction of an Association Graph}

We present the usual constructive definition of an association graph for
\abb{GMP} (\ref{nlp:GMP:gmp}) defined on a $\mathfrak{p}$-closed search
space $\S{M}$.

For each element $(\vec{i}, \vec{r}) \in I_{XY}$, we check whether item
$\vec{i}$ from $X$ and item $\vec{r}$ from $Y$ can be associated to a vertex
or edge in an association graph. The morphisms from the search space $\S{M}$
determine the association rule via its describing property $\mathfrak{p}$.
We associate item $\vec{i}$ from $X$ with item $\vec{r}$ from $Y$ if they are
$\mathfrak{p}$-similar. Thus, an \emph{association graph} $Z = X \diamond Y$
is a weighted graph with vertex and edge set
\begin{align*}
  V\big(Z\big) &= \Big\{ir \,:\, (i,i)
  \sim_{\mathfrak{p}} (r,r)\Big\}\subseteq V(X) \times V(Y) \\
  E\big(Z\big) &= \Big\{(ir, js) \,:\, (i,j)\sim_{\mathfrak{p}}(r,s)
  \Big\}\subseteq V(Z) \times V(Z).
\end{align*}
The matrix $\vec{Z} = (z_{irjs})$ with
\[
z_{irjs} = \begin{case}
  \epsilon & (ij, rs) \in \widebar{E}(Z)\\
  \kappa_{ijrs} & \mbox{otherwise}
\end{case}
\]
assigns weights to the vertices and edges of $Z$. Note that an association
graph assigns real-valued weights to both, its vertices \emph{and} edges.
This is in contrast to standard association graph formulations, where $Z$ is
either unweighted (constant weights) or weights are assigned to vertices
only.

\proofskip

\subsection{A Necessary and Sufficient Condition}

In this subsection, we show that $\mathfrak{p}$-closure of $\S{M}$ is a
necessary and sufficient condition for the desired equivalence between graph
matching and clique search in an association graph. Note that equivalence
means that there is a bijective mapping $\Phi:\S{C}_{Z} \rightarrow \S{M}$
with $\omega(C) = f\big(\Phi(C)\big)$ for all cliques $C \in \S{C}_{Z}$. The
mapping $\Phi$ then induces a one-to-one correspondence between the maximum
(maximal) cliques in $Z$ and the global (local) optimal solutions from
$\S{M}$.

\begin{theorem}\label{theorem:MWCP:gmp2mwcp}
  Let $X$ and $Y$ be graphs. Then the \abb{GMP} (\ref{nlp:GMP:gmp}) of $X$
  and $Y$ is equivalent to the \abb{MWCP} in a $\kappa$-association graph $Z
  = X\diamond Y$ if, and only if, there is a property $\mathfrak{p}$ on
  $I_{XY}$ such that the search space $\S{M}$ of (\ref{nlp:GMP:gmp}) is
  the set of all $\mathfrak{p}$-morphisms from $\S{M}_{XY}$.
\end{theorem}

\Proof The $\Rightarrow$ - direction is trivial. Suppose that \abb{GMP}
(\ref{nlp:GMP:gmp}) is equivalent to the \abb{MWCP} in $Z$.  Then we simply
define the property $\mathfrak{p}$ with $\vec{i}\sim_{\mathfrak{p}} \vec{r}$
if there is a morphism $\phi\in \S{M}$ with $\vec{i}^{\phi} = \vec{r}$.

Now let us show the opposite direction. Suppose that there is a property
$\mathfrak{p}$ on $I_{XY}$ such that $\S{M}$ is the set of all
$\mathfrak{p}$-morphisms from $\S{M}_{XY}$. We want to show that there is a
bijection
\[
\Phi:\S{C}_{Z} \rightarrow \S{M}, \quad C \mapsto \phi_{C}
\]
such that $\omega(C) = f(\phi_{C}, X, Y)$ for all $C \in \S{C}_{Z}$. With
each clique $C \in \S{C}_{Z}$ we associate a partial morphism $\phi_{C}:V(X)
\rightarrow V(Y)$ such that $\phi_{C}(i) = r$ for all $(i,r) \in C$. We show
that $\phi_{C}$ is a feasible morphism from $\S{M}$. Let $i,j\in V(X)$ and
$r,s\in V(Y)$ be vertices with $\phi_{C}(i) = r$ and $ \phi_{C}(j) = s$. By
construction, $(i,r)$ and $(j,s)$ are members of clique $C$. Since $Z[C]$ is
complete, there is an edge incident with $(i,r)$ and $(j,s)$. Hence, $(i,r)
\sim_{\mathfrak{p}} (j,s)$ and, therefore, $\Gamma(\phi_{C}) \subseteq
\S{R}_{XY}^{\mathfrak{p}}$. Since $\S{M}$ is $\mathfrak{p}$-closed, we have
$\phi_{C}\in \S{M}$.

Similarly, with each morphism $\phi \in \S{M}$ we associate a subset
$C_{\phi}$ of $V(Z)$ with
\[
(i,j) \in C_{\phi} \;\Leftrightarrow\; \phi(i) = j \mbox{ or } \phi(j) = i.
\]
Since $\phi$ is a feasible $\mathfrak{p}$-morphism, $C_{\phi}$ is a clique
in $Z$. It is straightforward to show that both associations give rise to
well-defined mappings
\begin{align*}
  \Phi:\S{C}_{Z} \rightarrow \S{M},& \quad C \mapsto \phi_{C}\\
  \Psi:\S{M} \rightarrow \S{C}_{Z} ,& \quad \phi \mapsto C_{\phi}.
\end{align*}
From $\Phi\circ\Psi = \id$ on $\S{M}$ and $\Psi\circ\Phi= \id$ on
$\S{C}_{Z}$, it follows that $\Phi$ is bijective. Finally, the assertion
follows from
\[
\omega(C) = \sum_{ir,js \in C} z_{irjs} = \sum_{\vec{i} \in \S{D}(\phi_{C})}
\kappa_{\vec{i}\vec{i}^{\phi}} = f(\phi_{C}, X, Y).
\]
\qed

\subsection{Implications of Theorem 1}

Theorem \ref{theorem:MWCP:gmp2mwcp} has the following implications:
\begin{itemize}
\item Equivalence proofs now reduce to showing that the search space $\S{M}$
  of a \abb{GMP} is $\mathfrak{p}$-closed for some property $\mathfrak{p}$.
\item Theorem \ref{theorem:MWCP:gmp2mwcp} is the starting point for generic
  continuous solutions to \abb{GMP}s defined on $\mathfrak{p}$-closed search
  spaces. An equivalent continuous formulation of the \abb{MWCP} is given in
  \cite{bjjThesis}.
\end{itemize}

\section{Application of Theorem 1}

The aim of this section is threefold: First, we show that common graph
matching problems satisfy the necessary and sufficient condition. Second, we
want to illustrate how the necessary and sufficient condition simplifies
equivalence proofs. Third, we show the equivalence relationship for further
examples, such as the graph edit distance, for which the desired
relationship has been unproven.  

\subsection{Simple Exact Graph Matching Problems}\label{subsec:theorem1:exact}

In this subsection, we consider the problems listed in Table \ref{tab:EGMP}.
For the first two types of problems equivalence to clique search is
well-known. Homomorphic \abb{GMP} of type 3 in Table \ref{tab:EGMP} have
been considered in a slight variation for attributed trees \cite{Bartoli00}.

\begin{table}
\label{tab:EGMP}
\centering
\caption{Examples of standard graph matching problems.}
\begin{tabular}{cll}
\hline
\\[-2ex]
Type & Graph Matching Problem & $\S{M} \subseteq \S{M}_{XY}$\\
\hline
\\[-2ex]
1 & Maximum Common Subgraph Problem &  partial subgraph-morphisms \\
&$\rightarrow\;$ \textbf{Special case:} Subgraph Isomorphism Problem &  total subgraph-morphisms \\
\hline
\\[-2ex]
2 & Maximum Common Induced Subgraph Problem & partial isomorphisms \\
&$\rightarrow\;$ \textbf{Special case:} Induced Subgraph Isomorphism Problem &  total isomorphisms to subgraph of $Y$\\
&$\rightarrow\;$ \textbf{Special case:} Graph Isomorphism Problem & total isomorphisms\\
\hline
\\[-2ex]
3 & Maximum Common Homomorphic Subgraph Problem & partial homomorphisms \\
&$\rightarrow\;$ \textbf{Special case:} Subgraph Homomorphism Problem & total homomorphisms to subgraph of $Y$\\[0.5ex]
\hline
\end{tabular}
\end{table}

\introskip

An \emph{exact graph matching problem} is a \abb{GMP} (\ref{nlp:GMP:gmp}),
where the compatibility values of the matching objective are of the general
form
\begin{equation}\label{eq:GMP:Exact_GMP_kappa2}
\kappa_{\vec{i}\vec{j}} =
\begin{case}
  \alpha_{V}& \vec{x}_{\vec{i}} = \vec{y}_{\vec{j}}, \; \vec{i}\in V(X),\;
  \vec{j}\in V(Y)\\
  \alpha_{E}& \vec{x}_{\vec{i}} = \vec{y}_{\vec{j}}, \; \vec{i} \in E(X),\,
  \vec{j} \in E(Y)\\
  \alpha_{\widebar{E}}& \vec{i} \in \widebar{E}(X),\,\vec{j} \in
  \widebar{E}(Y)\\
  0 & \mbox{otherwise}
\end{case}
\end{equation}
for all items $\vec{i}$ of $X$ and $\vec{j}$ of $Y$. We require that the
parameters $\alpha_{V}$, $\alpha_{E}$, and $\alpha_{\widebar{E}}$ are
nonnegative such that $\alpha_{V}+\alpha_{E}+\alpha_{\widebar{E}} > 0$.
Thus, exact matching problems only credit exact correspondences between
items of $X$ and $Y$. Standard formulations of the problems listed in Table
\ref{tab:EGMP} aim at maximizing the cardinality of vertices of the common
substructure. Thus, we may set $\alpha_{V} = 1$ and $\alpha_{E} =
\alpha_{\widebar{E}} = 0$. If we want to maximize the cardinality of edges
of the common substructure, we may set $\alpha_{E} = 1$ and $\alpha_{V} =
\alpha_{\widebar{E}} = 0$. Other choices of the parameters $\alpha_{V}$,
$\alpha_{E}$, and $\alpha_{\widebar{E}}$ reflect the importance of an exact
association between corresponding items.

Different types of exact graph matching problems differ from the definition
of the search space $\S{M}$ as indicated in Table \ref{tab:EGMP}. Note
that the special cases considered here are \abb{GMP}s constrained over
subsets of total morphisms. Clearly, subsets of total morphisms are not
$\mathfrak{p}$-closed. Hence, in a strict sense, the special cases are not
equivalent to clique search. To argue consistently, we regard a special case
as an instance of the corresponding generic case, where we make an
additional decision based on the optimal solution. For example, we think of
the isomorphism problem as a \abb{MCISP}, where we decide that both graphs
under consideration are isomorphic if they have the same cardinality of
vertices as a maximum clique in a derived association graph. Using this
convention, we can show that the examples in Table \ref{tab:EGMP} are
equivalent to clique search in an association graph.

\begin{corollary}\label{cor:MWCP:gmp2mwcp:gmp2mwcp:examples1}
  Consider the \abb{GMP} (\ref{nlp:GMP:gmp}) with compatibility values of
  the form (\ref{eq:GMP:Exact_GMP_kappa2}). Then the problems listed in Table
  \ref{tab:EGMP} are equivalent to the \abb{MWCP} in an association graph.
\end{corollary}

\proof From Proposition \ref{prop:MWCP:gmp2mwcp:p-morphisms} follows that
the sets of partial subgraph-, iso-, and homomorphisms are
$\mathfrak{p}$-closed. For problems, which are maximized over total
morphisms, we may relax $\S{M}$ to partial morphisms without affecting the
optimal solutions of (\ref{nlp:GMP:gmp}), since $\alpha_{V}$, $\alpha_{E}$,
and $\alpha_{\widebar{E}}$ are nonnegative with
$\alpha_{V}+\alpha_{E}+\alpha_{\widebar{E}} > 0$. The assertion follows from
Theorem \ref{theorem:MWCP:gmp2mwcp}.  \qed

\subsection{Inexact Graph Matching Problems}\label{subsec:theorem1:inexact}

In this subsection, we consider inexact graph matching problems. Again let
$X$ and $Y$ be graphs with adjacency matrices $\vec{X} = (x_{ij})$ and
$\vec{Y} = (y_{ij})$.

\bigskip

\noindent
\emph{Best Common Subgraph Problems}

The \emph{best common subgraph problem} is a \abb{GMP} (\ref{nlp:GMP:gmp}),
where the compatibility values of the matching objective are arbitrary real
values. The search space $\S{M}$ is either the set of all partial
morphisms or the set of all partial monomorphisms from $X$ to $Y$.

\begin{corollary}\label{cor:MWCP:gmp2mwcp:gmp2mwcp:examples2}
  Let $X$ and $Y$ be attributed graphs. Then the best common subgraph
  problem is equivalent to the \abb{MWCP} in $Z = X \diamond Y$.
\end{corollary}

\proof Proposition \ref{prop:MWCP:gmp2mwcp:p-morphisms} and Theorem
\ref{theorem:MWCP:gmp2mwcp}.  \qed

\bigskip

\noindent
\emph{Probabilistic Graph Matching Problem}

In probabilistic graph matching, a probability model is drawn to measure
compatibility between items. The aim is to then find a morphism from $X$ to
$Y$ that maximizes a global maximum a posteriori probability. To describe
the probabilistic graph matching problem in formal terms, we first introduce
a distinguished \emph{null color} $\epsilon_{V}$ for vertices not contained
in $\S{A}$. Next, we extend the model $Y$ by including an isolated vertex
with null color $\epsilon_{V}$. The \emph{probabilistic graph matching
  problem} is defined by
\begin{equation}\label{nlp:GMP:probabilistic_gmp}
\mbox{$
  \begin{array}{l@{\qquad}l}
    \mbox{maximize}& {\displaystyle f\big(\phi, X, Y\big) =
    P\big(\phi \vert \vec{X}, \vec{Y}\big)}\\
    \mbox{subject to}
    & {\displaystyle \phi \in \S{M}_{XY}},
  \end{array}
  $}
\end{equation}
where the matching objective $P\big(\phi \vert \vec{X}, \vec{Y}\big)$ is the
a posteriori probability of $\phi$ given the \emph{measurements} $\vec{X}$
and $\vec{Y}$.\footnote{In accordance with the terminology used in
  \cite{Wilson97} we refer to $\vec{X}$ and $\vec{Y}$ as the sets of
  measurements.} Applying Bayes Theorem, we obtain
\begin{equation}\label{eq:GMP:prob_objective}
P\big(\phi \vert \vec{X}, \vec{Y}\big) = \frac{p(\vec{X}, \vec{Y}
\vert
  \phi) P(\phi)}{p(\vec{X}, \vec{Y})},
\end{equation}
where $P(\phi)$ is the joint prior for $\phi$. The quantities $p(\vec{X},
\vec{Y} \vert \phi)$ and $p(\vec{X}, \vec{Y})$ are the conditional
measurement density and the probability density functions, respectively, for
the sets of measurements.

Probabilistic graph matching problems are usually solved by a Bayesian
inference scheme that does not require explicit calculation of compatibility
values in advance. Conceptually, there are compatibility values and a
probabilistic graph matching problem turns out to be a \abb{GMP}
(\ref{nlp:GMP:gmp}) over the set of all total morphisms, where we extend the
graph $Y$ by an isolated vertex with void attribute $\epsilon_{V}$.

\begin{corollary}\label{cor:MWCP:gmp2mwcp:gmp2mwcp:examples3}
  The probabilistic graph matching problem of $X$ and $Y$ is equivalent to
  the \abb{MWCP} in $Z = X \diamond Y'$, where $Y'$ extends $Y$ by an
  isolated vertex with void attribute $\epsilon_{V}$.
\end{corollary}

\proof Proposition \ref{prop:MWCP:gmp2mwcp:p-morphisms} and Theorem
\ref{theorem:MWCP:gmp2mwcp}.  \qed

\bigskip

\noindent
\emph{Graph Edit Distance Problem}

The concept of graph edit distance generalizes the Levenshtein edit distance
originally defined for strings \cite{Levenshtein66}. The graph edit-distance
is defined as the minimum cost over all sequences of basic edit operations
that transform $X$ into $Y$. Following common use, the set of basic edit
operations are \emph{substitution}, \emph{insertion}, and \emph{deletion} of
items. Different cost functions can be assigned to each edit operation.

The sequences of edit operations that make $X$ and $Y$ isomorphic can be
identified with the partial monomorphisms $\phi : V(X) \rightarrow V(Y)$.
Each partial monomorphism $\phi$ induces a bijection $\phi:\S{D}(\phi)
\mapsto \S{R}(\phi)$ from the domain $\S{D}(\phi)$ to the range
$\S{R}(\phi)$ of $\phi$. In terms of $\phi$, the edit operations have the
following form
\begin{itemize}
\item \emph{Substitution}: An item $\vec{i}$ from $\S{D}(\phi)$ is
  \emph{substituted} by item $\vec{i}^{\phi}$ from $\S{R}(\phi)$.
\item \emph{Deletion}: Items $\vec{i}$ of $\overline{\S{D}}(\phi) = I(X)
  \setminus \S{D}(\phi)$ are \emph{deleted} from $X$.
\item \emph{Insertion}: Items $\vec{j}$ from $\overline{\S{R}}(\phi) = I(Y)
  \setminus \S{R}(\phi)$ are \emph{inserted} into $Y$.
\end{itemize}
If $x_{\vec{i}} = y_{\vec{i}^{\phi}}$, the substitution $\vec{i}$ by
$\vec{i}^{\phi}$ is called \emph{identical substitution}. The cost of a
partial monomorphism $\phi$ is then defined by
\begin{equation}\label{eq:GMP:Cost:MGEP}
  f\big(\phi, X, Y\big) = \sum_{\vec{i}\,\in\,\overline{\S{D}}(\phi)}
  C_{del}(\vec{i}) +
  \sum_{\vec{j}\,\in\,\overline{\S{R}}(\phi)} C_{ins}(\vec{j}) \\
  + \sum_{\vec{i}\,\in\,\S{D}({\phi})} C_{sub}\left(\vec{i}, \vec{i}^{\phi}
  \right),
\end{equation}
where $C_{del}(\vec{i})$ is the cost of deleting item $\vec{i}$ of $X$,
$C_{ins}(\vec{i})$ is the cost of inserting an item $\vec{i}$ into $Y$, and
$C_{sub}(\vec{i}, \vec{j})$ is the cost of substituting an item $\vec{i}$
from $X$ by an item $\vec{j}$ of $Y$. We assume that all costs are
nonnegative.

The \emph{graph edit distance problem} is then of the form
\begin{equation}\label{nlp:GMP:MGEP}
\mbox{$
  \begin{array}{l@{\qquad}l}
    \mbox{minimize}& {\displaystyle f\big(\phi, X, Y\big)}
    \\[1ex]
    \mbox{subject to}
    & {\displaystyle \phi \in \S{M}}
  \end{array}
  $}
\end{equation}
where the matching objective $f$ is defined as in (\ref{eq:GMP:Cost:MGEP})
and $\S{M}$ is the subset of all partial monomorphisms from $X$ to $Y$. The
constrained global maximum of $-f$ is the \emph{graph edit distance} of $X$
and $Y$.

To show equivalence between the graph edit distance problem and clique
search, we transform problem (\ref{nlp:GMP:MGEP}) to our standard form of a
\abb{GMP} as given in (\ref{nlp:GMP:gmp}). First, we expand the set $\S{A}$
of attributes to $\S{A}' = \S{A} \cup \{\texttt{d}\}$ by including the
distinguished symbol \texttt{d}. We call items with attribute \texttt{d}
\emph{dummy items}.  Next, we expand $X$ and $Y$ by adding dummy vertices.
Suppose that $X$ and $Y$ are of order $|X| = n$ and $|Y| = m$, respectively.
Insert $m$ dummy vertices into $X$ and $n$ dummy vertices into $Y$.
Connected each dummy vertex with all the other (original and dummy) vertices
by dummy edges. Let $X'$ and $Y'$ be the resulting expanded graphs. We call
$X'$ and $Y'$ the \emph{dummy extensions} of $X$ and $Y$. Finally, we define
an appropriate compatibility function $\kappa$. To this end, we first
introduce some auxiliary notations. By $V(X,a)$ we denote the subset of all
vertices of $X$ that have attribute $a$. Furthermore, let $C'_{sub}$ be a
nonnegative real-valued function on $I(X')\times I(Y')$ such that
\[
C'_{sub}(\vec{i},\vec{j}) = \left\{\begin{array}{c@{\quad : \quad}l}
    C_{sub}(\vec{i},\vec{j}) & \vec{i}\in I(X), \, \vec{j}\in I(Y)\\
    0 & \mbox{otherwise}
\end{array} \right. .
\]
Now we consider the following compatibility values of $X'$ and $Y'$
\begin{equation}\label{eq:MWCP:Edit_GMP_kappa}
\kappa_{\vec{ij}} = \left\{\begin{array}{l@{\quad : \quad}l}
    -C_{del}(\vec{i}) & \vec{i}\in V(X), \, \vec{j} \in V(Y', \texttt{d}) \\
    -C_{ins}(\vec{j}) & \vec{i}\in V(X', \texttt{d}), \, \vec{j} \in V(Y) \\
    -C'_{sub}(\vec{i},\vec{j}) & \mbox{otherwise}
  \end{array} \right.
\end{equation}
for all items $\vec{i}$ of $X'$ and $\vec{j}$ of $Y'$. Then the \abb{GMP}
over the set of all total monomorphisms from $X'$ to $Y'$ is equivalent to
the graph edit distance problem (\ref{nlp:GMP:MGEP}). Note that we consider
negative costs as compatibility values to turn the minimization problem
(\ref{nlp:GMP:MGEP}) into a maximization problem as in (\ref{nlp:GMP:gmp}).

\begin{corollary}\label{cor:MWCP:gmp2mwcp:gmp2mwcp:examples4}
  Let $X$ and $Y$ be attributed graphs. Then the graph edit distance problem
  is equivalent to the \abb{MWCP} in $Z = X' \diamond Y'$, where $X'$ and
  $Y'$ are the dummy extensions of $X$ and $Y$.
\end{corollary}

\proof Proposition \ref{prop:MWCP:gmp2mwcp:p-morphisms} and Theorem
\ref{theorem:MWCP:gmp2mwcp}.  \qed

\proofskip

\noindent
Solving the graph edit distance by clique search in $Z = X' \diamond Y'$ is
impractical, because the dummy extensions double the total number of
vertices of both original graphs. In a practical implementation, it is
sufficient to expand each of both graphs $X$ and $Y$ by adding only one
dummy vertex as described above. This leads to a formulation of the graph
matching problem constrained over relations $\phi \subseteq V(X) \times
V(Y)$ rather than partial morphisms. Though the theory developed in this
contribution also holds for relations, we do not consider the more general
case for the sake of clarity.

\section{Conclusion}\label{sec:conclusion}

We presented a necessary and sufficient condition (C) for graph matching to
be equivalent to the maximum weight clique problem in an association graph.
The implications of this result are as follows: first, equivalence proofs
now reduce to showing that condition (C) holds; second, the condition (C) is
applicable to a broad range of common graph matching problems; third,
generic continuous solutions can now be applied to all graph matching
problems that satisfy (C).

We showed that the graph edit distance problem, probabilistic graph
matching, the best common graph matching, and the maximum common
 homomorphic subgraph problem are equivalent to clique search in a derived association
graph.

One limitation with this framework motivates further research: The maximum
weight clique problem, where weights are assigned to both vertices and edges
has been rarely studied in the literature. Hence, it is conducive to
develop special continuous formulations and solutions to the maximum weight
clique problem in order to obtain a generic graph matching solver.


\bibliographystyle{plain} 
\bibliography{ref,bjj}

\begin{thebibliography}{10}

\bibitem{Ambler73}
A.P. Ambler, H.G. Barrow, C.M. Brown, R.M. Burstall, and R.~J. Popplestone.
\newblock A versatile computer-controlled assembly system.
\newblock In {\em International Joint Conference on Artificial Intelligence},
  pages 298--307. Stanford University, California, 1973.

\bibitem{Barrow76}
H.~Barrow and R.~Burstall.
\newblock Subgraph isomorphism, matching relational structures and maximal
  cliques.
\newblock {\em Information Processing Letters}, 4:83--84, 1976.

\bibitem{Bartoli00}
M.~Bartoli, M.~Pelillo, K.~Siddiqi, and S.W. Zucker.
\newblock Attributed tree homomorphism using association graphs.
\newblock In {\em Proceedings of the IEEE International Conference on Pattern
  Recognition}, pages 2133--2136, 2000.

\bibitem{Bunke97}
H.~Bunke.
\newblock On a relation between graph edit distance and maximum common
  subgraph.
\newblock {\em Pattern Recognition Letters}, 18(8):689--694, 1997.

\bibitem{Chen96}
C.-W.K. Chen and D.Y.Y. Yun.
\newblock Toward solving maximal overlap set problems.
\newblock Technical Report TR-LIPSC\&Y96a, Laboratory of Intelligent and
  Parallel Systems, University of Hawaii, 1996.

\bibitem{Chen98}
C.-W.K. Chen and D.Y.Y. Yun.
\newblock Unifying graph-matching problem with a practical solution.
\newblock In {\em Proceedings of International Conference on Systems, Signals,
  Control, Computers}, 1998.

\bibitem{Garey79}
M.~Garey and D.~Johnson.
\newblock {\em Computers and Intractability: A Guide to the Theory of
  NP-Completeness}.
\newblock W.H. Freeman and Company, New York, 1979.

\bibitem{Gentner83}
D.~Gentner.
\newblock Structure-mapping: A theoretical framework for analogy.
\newblock {\em Cognitive Science}, 7(2):155--170, 1983.

\bibitem{Goldstone94}
R.L. Goldstone.
\newblock Similarity, interactive activation, and mapping.
\newblock {\em Journal of Experimental Psychology: Learning, Memory, and
  Cognition}, 20(3):3--28, 1994.

\bibitem{bjjThesis}
B.J. Jain.
\newblock {\em Structural Neural Learning Machines}.
\newblock PhD Thesis, Berlin University of Technology, 2005.

\bibitem{Johnson88}
M.~Johnson.
\newblock Some constrains on embodied analogical understanding.
\newblock In D.H. Helman, editor, {\em Analogical Reasoning. Perspectives of
  Artificial Intelligence, Cognitive Science, and Philosophy}. Kluwer Academic
  Publishers, 1988.

\bibitem{Kann92}
V.~Kann.
\newblock On the approximability of {NP}-complete optimization problems.
\newblock Master's thesis, Dept.~of Numerical Analysis and Computing Science,
  Royal Institute of Technology, Stockholm, 1992.

\bibitem{Levenshtein66}
V.~Levenshtein.
\newblock Binary codes capable of correcting deletions, insertions and
  reversals.
\newblock {\em Soviet Physics-Doklady}, 10:707--710, 1966.

\bibitem{Levi72}
G.~Levi.
\newblock A note on the derivation of maximal common subgraphs of two directed
  or undirected graphs.
\newblock {\em Calcolo}, 9:341--352, 1972.

\bibitem{Pelillo99b}
M.~Pelillo.
\newblock Replicator equations, maximal cliques, and graph isomorphism.
\newblock {\em Neural Computation}, 11(8):1933--1955, 1999.

\bibitem{Pelillo02}
M.~Pelillo.
\newblock Matching free trees, maximal cliques, and monotone game dynamics.
\newblock {\em IEEE Transactions on Pattern Analysis and Machine Intelligence},
  24(11):1535--1541, 2002.

\bibitem{Pelillo99d}
M.~Pelillo, K.~Siddiqi, and S.W. Zucker.
\newblock Attributed tree matching and maximum weight cliques.
\newblock In {\em Proc. ICIAP'99-10th Int. Conf. on Image Analysis and
  Processing}, pages 1154--1159. IEEE Computer Society Press, 1999.

\bibitem{Pelillo99a}
M.~Pelillo, K.~Siddiqi, and S.W. Zucker.
\newblock Matching hierarchical structures using association graphs.
\newblock {\em IEEE Transactions on Pattern Analysis and Machine Intelligence},
  21(11):1105--1120, 1999.

\bibitem{Raymond02}
J.W. Raymond, E.J. Gardiner, and P.~Willett.
\newblock {RASCAL}: Calculation of graph similarity using maximum common edge
  subgraphs.
\newblock {\em Computer Journal}, 45(6):631--644, 2002.

\bibitem{Schaedler99a}
K.~Sch\"adler and F.~Wysotzki.
\newblock Comparing structures using a {H}opfield-style neural network.
\newblock {\em Applied Intelligence}, 11:15--30, 1999.

\bibitem{Wilson97}
R.C. Wilson and E.R. Hancock.
\newblock Structural matching by discrete relaxation.
\newblock {\em IEEE Transactions on Pattern Analysis and Machine Intelligence},
  19(6):634--648, 1997.

\end{thebibliography}

\end{document}